\documentclass[sigconf]{acmart}
\usepackage{tikz}
\usepackage{xcolor}
\usetikzlibrary{arrows.meta, positioning, quotes}

\AtBeginDocument{%
  \providecommand\BibTeX{{%
    \normalfont B\kern-0.5em{\scshape i\kern-0.25em b}\kern-0.8em\TeX}}}

\setcopyright{rightsretained}

\begin{document}
\copyrightyear{2021} 
\acmYear{2021} 

\acmConference[AIES '21]{Proceedings of the 2021 AAAI/ACM Conference on AI, Ethics, and Society}{May 19--21, 2021}{Virtual Event, USA}
\acmBooktitle{Proceedings of the 2021 AAAI/ACM Conference on AI, Ethics, and Society (AIES '21), May 19--21, 2021, Virtual Event, USA}
\acmDOI{10.1145/3461702.3462467}
\acmISBN{978-1-4503-8473-5/21/05}
\fancyhead{}

\title{Causality in Neural Networks - An Extended Abstract}

\author{Abbavaram Gowtham Reddy}
\email{cs19resch11002@iith.ac.in}
\affiliation{%
  \institution{Indian Institute of Technology Hyderabad}
  \streetaddress{Kandi}
  \city{Hyderabad}
  \state{Telangana}
  \country{India}
  \postcode{502285}
}

\renewcommand{\shortauthors}{Abbavaram Gowtham Reddy}

\begin{abstract}
 Causal reasoning is the main learning and explanation tool used by humans. AI systems should possess causal reasoning capabilities to be deployed in the real world with trust and reliability. Introducing the ideas of causality to machine learning helps in providing better learning and explainable models. Explainability, causal disentanglement are some important aspects of any machine learning model. Causal explanations are required to believe in a model's decision and causal disentanglement learning is important for transfer learning applications. We exploit the ideas of causality to be used in deep learning models to achieve better and causally explainable models that are useful in fairness, disentangled representation, etc.
\end{abstract}



\maketitle
\section{Need for Explainable AI}
There are a `wide variety of application domains where Deep Learning models have shown human level performance or even better in terms of accuracy or decision making \cite{deng2014deep, sadowski2014searching, Nelson2017StockMP}. But Deep Learning models are considered to be \textit{black-box} models because the internal workings are not clearly known. This leaves us with the questions of reliability and trustworthiness of such models. Explanation is an essential quality of any machine learning model to be used in real-world applications especially in critical domains such as medicine, defence, space science etc. Along with the accuracy, explanation should also be an evaluating criteria of machine learning models asking that “What is your decision(accuracy) and why is that decision(explanation)?”

For human understandable explanations, we need to have machine learning models that have high-level knowledge representation and reasoning capabilities. Several mathematical techniques such as causality, logic, and probability will help us in providing human understandable explanations. Broadly we can categorize the explanation based models into perturbation based vs human evaluated explanation models, local vs global explanation models, post-hoc vs inbuilt explanation models, etc.

\section{Open challenges}
Neural Networks are good in getting high accuracy but not good at generating explanations for their decisions. Simple models such as linear regression, decision tree are good at providing the basis for their explanations but they are not good at achieving state of the art performance due to limited modelling capabilities. The trade-off between accuracy and explanation achieved by different models can be visualised in classical accuracy vs explainability graph \cite{xaitutorial}.
Another problem with regular machine learning models is that they learn correlations in the data rather than causal relationships. Making neural networks to learn causal relationships(which are essential for human understandable explanations) rather than correlations is still an open problem. Fairness is another important application domain where a machine learning model should not rely on the sensitive/protected attributes(e.g., gender, race) to arrive at a decision. Few attempts were made in this area \cite{joo2020gender, kusner2017counterfactual} and still there is a good scope in improving the models. Incorporating causal reasoning capabilities in neural networks in terms of regularization is also an open problem. Learning independent causal mechanisms\cite{Peters2017} from observational data that are disentangled according to an underlying causal process is important for transfer learning tasks and thus important to achieve general intelligence.

\section{Current Work}
\begin{figure}[H]
    \centering
      \includegraphics[width=0.22\textwidth]{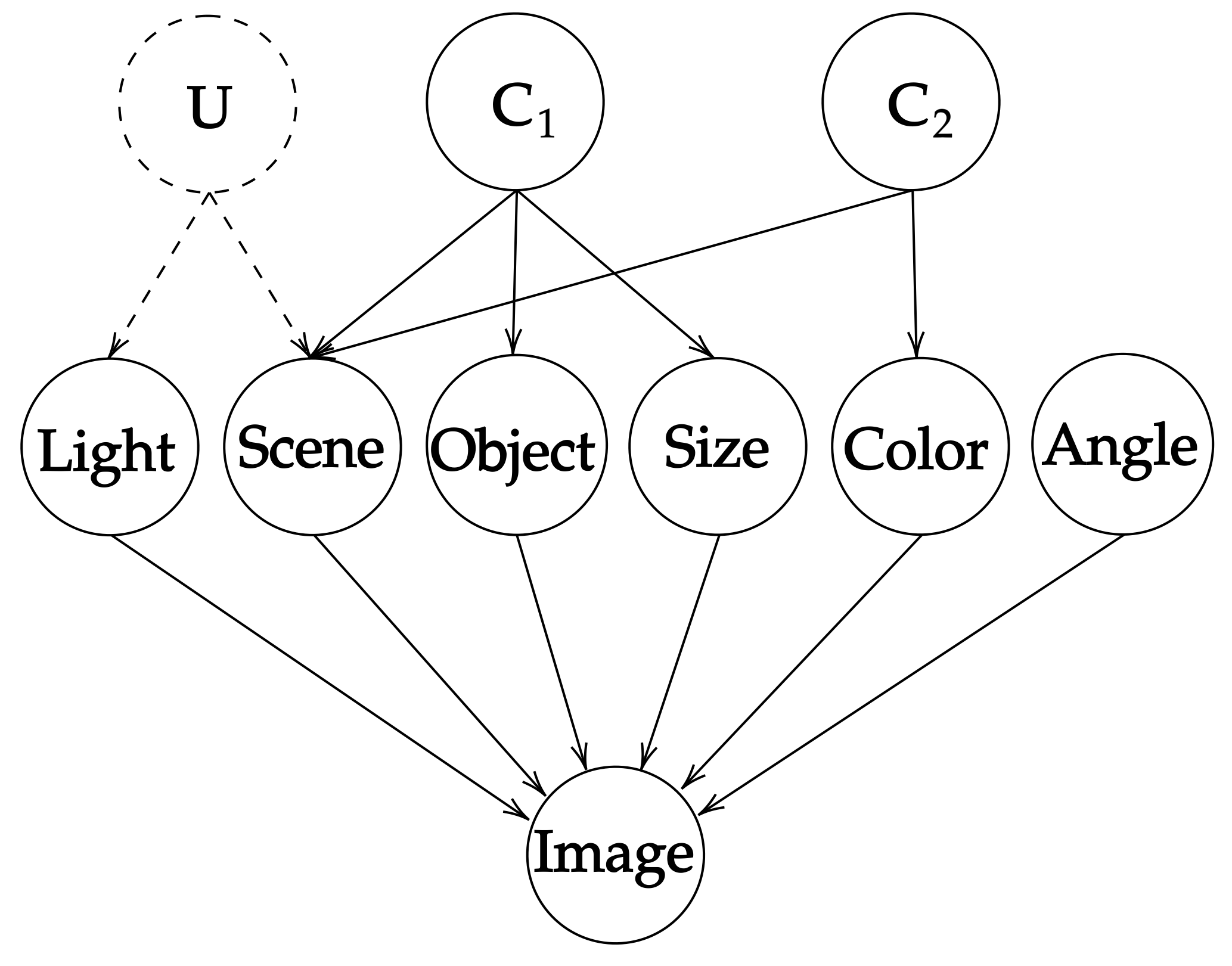}
      \caption{Data generating DAG with both observed($C_1,C_2$) and unobserve($U$) confounders.}
    \label{fig:datagenerator_confounding}
\end{figure}

Inspiring from the above open challenges, we currently take the following research directions.
\paragraph{\textbf{CANDLE}}
We develop an image dataset for \textbf{C}ausal \textbf{AN}alysis in \textbf{D}isentang\textbf{L}ed r\textbf{E}presentations(CANDLE), a realistic dataset generated following the causal graph shown in Figure \ref{fig:datagenerator_confounding}. Data generating mechanism has both observed and unobserved confounders. Confounding here can be thought of as few objects appear in a specific color in a specific background which is very natural in real-world. Our interest is to ask counterfactual questions on the dataset to evaluate methods that learn the representations. Along with the dataset, we provide different supervisions that can be used by various methods to identify and remove confounding effects.  The dataset is generated with the help of Blender~\cite{blender}, a free and open-source 3D computer graphics software, which allows for manipulating the background HDRI images and adding foreground elements that inherit the natural light of the background. Foreground elements naturally cast shadows to interact with the background. This greatly increases the realism of the dataset while allowing for it to remain simulated.
We also propose two metrics to measure the level of disentanglement achieved by a model trained under confounding dataset. Evaluation of disentangled representation learners under confounding effects is useful for deciding which model to use when the dataset is confounded. We propose two evaluation metrics called: \textit{unconfoundedness} and \textit{counterfactual generativeness} metrics that capture the entire causal process of data encoding and data generation of a VAE \cite{kingma2013auto} based model.
We define \textit{unconfoundedness}($UC$) measure as
\begin{equation}
\label{eq:uc}
    UC \coloneqq 1 - \mathbb{E}_{x\sim \mathbb{D}}\big [\frac{1}{K} \sum_{i\ne j} \frac{|Z_i^x \cap Z_j^x|}{|Z_i^x \cup Z_j^x|}\big ]
\end{equation}
Where $K= \binom{N}{2}$, the number of pairs of generative factors. $Z_i^x$ is the set of latent dimensions that encode information about generative factor $i$ in an image $x$ taken from the dataset $\mathbb{D}$. We take the \textit{jaccard similarity} between each pair of encodings of generative factors to measure the level of disentanglement. We propose \textit{counterfactual generativeness}($CG$) by defining the average causal effect(ACE) of latents $Z_i/Z_{\setminus i}$ on the generated counterfactual image $X^{cf}_{i}/X^{cf}_{\setminus i}$ as
\begin{equation}
    CG \coloneqq \mathbb{E}_i \big[| ACE^{X^{cf}_{i}}_{Z_{i}} - ACE^{X^{cf}_{\setminus i}}_{Z_{\setminus i}}|\big]
\end{equation}
$X^{cf}_i$ represents the counterfactual image generated when latent factors $Z_i$ are set to some interventional values.
Where we define $ ACE^{X^{cf}_{i}}_{Z_{i}}$ to be the difference in prediction probability of generative factors $i$ given the counterfactual image $x^{cf}_i$. We empirically analyze existing models on synthetic, dSprites\cite{dsprites17}, MPI3D-Toy\cite{gondal2019transfer}, and CANDLE datasets.
\paragraph{\textbf{CREDO}}
Instead of relying on posthoc explainations \cite{lime,selvaraju2017grad, chattopadhyay2019neural}, we seek to develop explainable neural network models by introducing causal regularization as an additional objective along with empirical risk minimization. We propose a model: \textbf{C}ausal \textbf{RE}gularization through \textbf{DO}main priors(CREDO) that can incorporate any given causal priors into the network in terms of both (controlled) direct and (natural) total causal effects\cite{pearl2001direct}. Very few attempts \cite{javed2020learning, sen2018supervising} were made in this direction. 

Using the idea of looking at a neural network $\mathcal{F}$ as SCM\cite{chattopadhyay2019neural}, we differentiate the direct and total causal effects learned by $\mathcal{F}$ and explicitly regularize the model to learn the prior function given at the time of training. If we look at $\mathcal{F}$ as an SCM that learns structural equations while training, we identified that the $n^{th}$ partial derivative of Average Controlled Direct Effect $(ACDE)$ of $i^{th}$ input feature $x_i$ on $\mathcal{F}$ is equal to the mean of $n^{th}$ partial derivatives of $\mathcal{F}$ w.r.t. $x_i$, that is: $\frac{\partial^n ACDE^{\mathcal{F}}_{x_i}}{\partial x^n_i} = \mathbb{E}\left[\frac{\partial^n \mathcal{F}}{\partial x^n_i}\right]$. So it is enough to train a model that matches the gradient of causal prior functions to the gradients of neural networks to achieve desired causal effect in neural networks. We also have a similar result that relates Average Total Causal Effect and the total gradient of the neural network. All of this is possible because we were able to look neural network as an SCM that relates inputs and outputs by marginalising out the hidden layers~\cite{chattopadhyay2019neural}.
\vspace{-2pt}
\paragraph{\textbf{Survey}}
We also work on providing a survey of recent techniques that exploit the ideas of causality in machine learning from various aspects like fairness, computer vision, disentanglement etc. Survey article explains the flow of ideas  
from causality and machine learning(e.g., counterfactual, explainability) to applications(e.g., fairness, domain generalization).
\vspace{-2pt}
\bibliographystyle{ACM-Reference-Format}
\bibliography{bibliography}
\end{document}